\documentclass[review]{elsarticle}

\usepackage{lineno,hyperref}
\usepackage[inline]{enumitem}
\usepackage{algorithm}
\usepackage{algorithmic}
\usepackage{amsmath}
\usepackage{amssymb}
\usepackage{bm}
\usepackage{caption}
\usepackage{setspace}
\usepackage{multirow}
\usepackage{tabularx,booktabs}
\usepackage[mathcal]{eucal}
\usepackage{graphicx}
\modulolinenumbers[5]

\journal{Journal of \LaTeX\ Templates}

\begin{document}

\begin{frontmatter}

\title{A Coevolutionary Approach to Experience-based Optimization}

\author[ustc]{Shengcai Liu}
\ead{liuscyyf@mail.ustc.edu.cn}
\author[ustc]{Ke Tang\corref{cor1}}
\ead{ketang@ustc.edu.cn}
\author[honda]{Bernhard Sendhoff}
\ead{bernhard.sendhoff@honda-ri.de}
\author[nust]{Xin Yao}
\ead{xiny@sustc.edu.cn}

\cortext[cor1]{Corresponding author}
\address[ustc]{USTC-Birmingham Joint Research Institute in Intelligent Computation
and Its Applications (UBRI), School of Computer Science and Technology,
University of Science and Technology of China, Hefei, Anhui, China}
\address[honda]{Honda Research Institute Europe GmbH, Offenbach 63073, Germany}
\address[nust]{Department of Computer Science and Engineering, Southern University of Science and Technology, Shenzhen, China}



\begin{abstract}
Solvers for hard optimization problems are usually built
to solve a set of problem instances (e.g., a problem class), rather
than a single one. It would be desirable if the performance of a
solver could be enhanced as it solves more problem instances,
i.e., gains more experience.
This paper studies how a solver could be enhanced based on its past problem-solving
experiences. Specifically, we focus on offline methods which can be
understood as a training process of solvers before deploying them.
Previous research arising from different aspects are first reviewed
in a unified context, termed experience-based optimization.
The existing methods mainly deal with the two issues in training,
i.e., training instance selection and adapting the solver to the training instances, in two independent phases, although these two issues are correlated
since the performance of the solver is dependent on the instances on which it is trained. Hence, a new method, dubbed LiangYi, is proposed to address these two issues simultaneously. LiangYi maintains a set of solvers and a set of instances. In the training process,
the two sets co-evolve and compete against each other, so that LiangYi can iteratively identify new training instances that are challenging for the current solver population and improve the current solvers. An instantiation of LiangYi on the Travelling Salesman Problem is presented. Empirical results on a test set containing 10000 instances showed that LiangYi could evolve solvers that perform significantly better than the solvers trained by other state-of-the-art training methods. Empirical studies on the behaviours of LiangYi also confirmed that it was able to continuously improve the solver through coevolution.
\end{abstract}


\end{frontmatter}

\section{Introduction}


Hard optimization problems (e.g., NP-hard problems) are ubiquitous in artificial intelligence (AI) research and real-world applications. To tackle them, numerous solvers have been proposed over the last few decades \cite{papadimitriou1982combinatorial}. In general, a solver is designed for a certain problem domain rather than a single instance, because when used in practice, it usually needs to solve many different instances belonging to that domain.
Many such solvers are heuristic methods, the performance (e.g., time complexity required to obtain the optimal solution) of which can hardly be rigorously proved. As a result, the development of these solvers typically involves repeatedly testing it against a number of problem instances and adjusting it based on the test results \cite{hoos2011automated}.
Given that both the design and the applications of a solver would involve many problem instances, a natural question is whether a solver could leverage on the experience acquired from solving previous problem instances to grow/enhance its capacity in solving new problem instances.
This simple intuition motivated the term
that is referred to as experience-based optimization (EBO) in our work.
EBO concerns designing mechanisms
that can improve the performance of a solver as it solves more and more problem instances.

The intuitions behind EBO are two-fold. First, any human expert in a specific domain starts as a novice and his/her path to an expert mainly relies on the gradual accumulation of problem-solving experience in this domain. Second, exploiting past experience to facilitate the solving of new problems, from a more technical point of view, concerns the generalization of past experience, which lies in the heart of AI research, particularly in machine learning. The past few decades have witnessed great progresses in machine learning, where most successes were achieved in building a learner that can correctly map an input signal (e.g., an image) to a predefined output (e.g., a label). It is interesting to ask whether similar idea could be developed to encompass more complex problems such as NP-hard optimization problems, which may introduce new challenges as the desired output will no longer be a label (or other types of variables), but a solution to the optimization problem.

EBO could offer three advantages in practice. First, it enables an automated process analogous to life-long learning and thus alleviate the tedious step-by-step fine-tuning or upgrade work that is now mostly done by human experts.
Second, as EBO methods improve the performance of the solver automatically, it would be able to better exploit high-performance computing facilities to generate and test much more problem instances than a domain expert can do manually, such that the risk of over-tuning the solver to a small set of problem instances can be reduced. Finally, the underlying properties of real-world hard optimization problem instances, even if they are from exactly the same problem class, may change over time.
Since in EBO solvers are dynamically updated when solving more and more problem instances,
they could better handle the changing world.

Similar to many machine learning techniques, EBO may run in two modes, i.e., the offline and online modes. For the offline mode, a set of problem instances is fed to an EBO method and the solver is updated after collecting the solutions it obtained on all the instances. For the online mode, problem instances are fed to an EBO method one at a time and the solver is updated immediately after solving an instance. In practice, the offline mode might be more important since a set of training instances are usually available when designing a solver. The online mode is more likely to occur after the solver is deployed in a real-world application. Even in this case, the offline mode could still be adopted since the solver could be updated until collecting more instances. In this paper, we focus on the offline mode as the first step to investigate EBO. Specifically, our work consists of three main parts, as summarized below:

\renewcommand\labelenumi{(\theenumi)}
\begin{enumerate}
\item A systematic overview of the key issues in EBO: In the literature, there have been several attempts to design mechanisms that enhance a solver based on past experience. For examples, transfer methods \cite{BoyanM00,feng2015memetic,santana2012structural}, as the name implies, transfer the useful information extracted from solved instances to unsolved problem instances. Automatic algorithm configuration \cite{hoos2011automated, hutter2009paramils}, portfolio-based algorithm selection \cite{rice1976algorithm, smith2009cross, kotthoff2014algorithm}, and automatic portfolio construction \cite{xu2010hydra, xu2011hydra, kadioglu2010isac, malitsky2012instance, tang2014population, tang2016negatively}
seek to identify better parameter settings, algorithm selectors, and portfolios of algorithms, respectively, based on historical data. Although all these methods are within the scope of EBO and thus relevant to one another, they were developed through independent paths and have never been discussed in a unified context. We first bring together the existing literature on the offline scenario of EBO, and review them under the unified umbrella of EBO, so as to make the key issues in EBO clearer.

\item A new offline training approach for EBO. A (and probably the most fundamental) form of offline EBO methods is to train the solvers with many problem instances, so as to obtain well-developed solvers before deployment.
This scenario involves at least two questions, i.e., where the training instances come from and how the solver is adapted (trained) to the training instances. These two issues were usually treated through two independent phases and seldom addressed simultaneously in the literature. We argue that they are inter-correlated and hence propose a coevolutionary framework, namely LiangYi, to address them as a whole.

\item A case study of LiangYi on the Travelling Salesman Problem (TSP). To assess the potential of LiangYi, a specific instantiation of it is implemented based on the Chained Lin-Kernighan (CLK) algorithm for the TSP. Empirical studies are conducted to compare LiangYi to other state-of-the-art methods, as well as to investigate the properties of LiangYi.

\end{enumerate}
The rest of this paper is structured as follows. Section 2 first gives a definition of the offline training in EBO and presents the key issues for describing the offline training methods, and then review the existing methods. Section 3 presents the approach LiangYi. Section 4 instantiates LiangYi on the TSP and reports the empirical results. Section 5 concludes the paper and outlines directions for future research.

\section{Offline Training Methods in EBO}
\label{offline}
Given an optimization task $T$ and a performance metric $m$,
the training in EBO
is defined as
improving a solver $s$
on optimization task $T$ with respect to
performance metric $m$ through experience $E$.
This definition borrows some basic concepts (i.e., $T$, $m$, $E$)
from the definition of machine
learning by Mitchell \cite{mitchell1997machine},
yet each of them has a concrete meaning here.
Specifically, the optimization task $T$ is conceptually an
instance set containing all the target instances to which the
solver is expected to be applied.
The performance metric $m$ is user-specified and it is often related to
the computational resources consumed by the solver
(such as runtime or memory) or the quality of the solution found.
Conceptually, the experience $E$ includes all possibly useful information which could be obtained from the solver s solving training instances, such as the state of $s$ (e.g., the parameter values of $s$), the solved instances and the corresponding solutions obtained by $s$ on them, the time needed by s to find these solutions, and the processes of s solving these instances (e.g., the search trajectory for a search based solver).
Although all these information could be useful for improving $s$, different training methods may focus on using different information, which we will discuss more in the next few sections.

To improve the solver $s$ on task $T$,
a training method must consider two things: How to use $s$ and the training instances to
produce useful experience and how to
exploit the experience so as to enhance $s$.
Generally, a solver $s$ is comprised of
multiple different parts (for example, a search-based solver
includes at least an initialization module and a search operator).
It is conceivable that
if $s$ is improved by training, some parts of $s$ are necessarily changed
during the training process.
Based on these analyses, we consider that
an offline training method consists of three essential parts:
\begin{itemize}
\item the form of the solver being trained;
\item the settings of the training instances;
\item the training algorithm that manipulates the solver and the training
instances to produce experience, and exploits the experience to improve the solver.
\end{itemize}
With this framework,
we can describe offline training methods in EBO in
a unified way.
In the combinatorial optimization field,
there have been various attempts by different communities
to obtain solvers through training.
The next few sections
review such research.

\subsection{Automatic Algorithm Configuration Methods} 
\label{sub:AAC}
The first class of methods
are automatic algorithm configuration ($AAC$)
methods \cite{hoos2011automated, hutter2009paramils}.
$AAC$ methods improve the solver (a parameterized algorithm)
on the optimization task by finding parameter values
under which the solver achieves high performance on
the target instances.
Specifically, $AAC$ methods adopt a two-stage strategy.
They first build a training set containing
the training instances
that are representative of the target instances,
and then run the training algorithms
to find high-performance parameter values
on the training set.
Due to the assumed similarity between the training instances and the target
instances, the found parameter configurations are expected to perform well
on the target instances as well.
A number of efficient methods have been developed in the field of $AAC$,
such as
CALIBRA \cite{adenso2006fine},
ParamILS \cite{hutter2009paramils},
GGA \cite{ansotegui2009gender},
SMAC \cite{hutter2011sequential} and
irace \cite{lopez2011irace}.
Using our framework presented previously,
$AAC$ methods can be expressed as follows:

\begin{itemize}
    \item The solver $s$ being trained is a parameterized algorithm.
        \footnote{Although there may be some significant differences between the parameterized algorithms
        (in the aspects such as the types of the parameters, or the number of the parameters) that different $AAC$ methods can handle, we choose to ignore these details because what we want to clarify here is which part of the solver is changed by the training, and the solver description, i.e., a parameterized algorithm, is enough for this purpose. Such a simplicity principle also applies in the reviews of other kinds of methods.}

    \item The efficacy of $AAC$ methods depend greatly on
		  the selection of the training instances, that is,
          the training instances should represent the
          target instances well so that the optimized performance
          on the training instances could be favourably transferred to the
          target instances. The usual practice in setting training instances for
          $AAC$ methods \cite{hutter2009paramils, ansotegui2009gender,hutter2011sequential,lopez2011irace} is that the training instances are directly selected from some benchmarks, or are randomly generated through some instance generators.
          Such practice is based on the assumption that the selected benchmarks and
          generators could represent the target scenarios to which the solver
          will be applied to.
          This assumption however has sparked some controversy
          \cite{hooker1995testing, smith2015generating},
          which we will discuss more in Section~\ref{LiangYisec}.
    \item Essentially, in the training process, $AAC$ methods would test different parameter configurations with the training set; therefore the experience $E$ produced is actually those tested parameter configurations and the corresponding test results.
    The way of exploiting $E$ is simple --- reserving the best-performing one.

          Different $AAC$ methods mainly differ in how they deal with the specific issues when producing $E$. Among these issues the most important ones include: Which parameter configurations should be evaluated, which instances should be used to evaluate a parameter configuration, how to reasonably compare two configurations, and when to terminate the evaluation of those poorly performing configurations. A detailed review of these aspects in this area is beyond the scope of this paper
          and one may refer to \cite{hoos2011automated, hutter2009paramils}
          for a more comprehensive treatment on the subject.

\end{itemize}

\subsection{Portfolio-based Algorithm Selection Methods} 
\label{sub:Algorithm Selection}
The second class of methods are
portfolio-based algorithm selection ($PAS$)
methods \cite{rice1976algorithm, smith2009cross, kotthoff2014algorithm}.
Although there are different
interpretations of this term "portfolio" in the literature,
we use it here to denote a solver that contains several
candidate algorithms and always selects one of them when solving
a problem instance.
To improve the solver (an algorithm portfolio),
unlike $AAC$ methods,
$PAS$ methods do not change the algorithms
that constitute the solver,
but build a selector
that can accurately select the best from the candidate algorithms
for each instance.
$PAS$ methods adopt the same two-stage strategy as $AAC$ methods,
except that the second-stage training algorithms in $PAS$ methods
are used to establish the selector.
Using our framework presented previously, $PAS$ methods can be expressed as follows:
\begin{itemize}
    \item The solver $s$ being trained is an algorithm portfolio.
    \item The training instance settings for $PAS$ methods are the same as $AAS$ methods.
    \item
    To build the algorithm selector, $PAS$ methods first gather performance data by running each candidate algorithm on the training instances.,
    and then build an algorithm selector based on the gathered data.
    The experience $E$ produced in the training is the performance data,
    and the exploitation of $E$ is carried out in this way:
    First suitable features that characterize problem instances are identified, and then the feature values of training instances are computed; Once each training instance is represented
    by a vector of feature values, the performance data ($E$) is transformed
    into a set of training data. Machine learning techniques are
    then used to learn from the training data a mapping from instance features to algorithms, which is exactly the algorithm selector.

    Different $PAS$ methods build different models to do the
    mapping (selection), such as regression
    models \cite{xu2008satzilla,
    gebser2011portfolio, leyton2009empirical, hutter2014algorithm, hoos2014claspfolio, lindauer2015autofolio}
    (so-called empirical performance models),
    classification models \cite{xu2012satzilla2012, kadioglu2011algorithm,malitsky2013boosting, malitsky2013algorithm, hoos2014claspfolio, lindauer2015autofolio}
    and ranking models \cite{kotthoff2012evaluation}.
    \footnote{Although many of the methods cited here are
    not originally proposed for combinatorial optimization problems,
    the ideas behind them
    are very general and apply to
    the combinatorial optimization problems as well.}
    For additional information one may refer to the survey of many approaches to algorithm
    selection from cross-disciplinary perspectives \cite{smith2009cross}
    and the constantly updated  survey \cite{kotthoff2014algorithm}  focusing on the contributions made in the area of combinatorial search problems.

\end{itemize}

\subsection{Automatic Portfolio Construction Methods} 
\label{sub:automatic_portfolio_construction}
The third class of methods are automatic portfolio construction ($APC$) methods, which seek to automatically build an algorithm portfolio
from scratch.
$APC$ methods \cite{xu2010hydra, xu2011hydra,kadioglu2010isac, malitsky2012instance} not only change the constituent algorithms in the portfolio, but also establishes an algorithm selector.
Another class of portfolio construction methods, called automatic parallel portfolio construction ($APPC$) methods \cite{lindauer2016automatic,
tang2014population}
differ from $APC$ methods in that they seek to construct
a \textit{parallel} algorithm portfolio that
runs all candidate algorithms in parallel when solving an instance.
In other words, $APPC$ methods also change the constituent algorithms in the portfolio,
but do not involve any algorithm selection.
Like $AAC$ methods and $PAS$ methods, both $APC$ methods and $APPC$ methods also adopt a two-stage strategy.
Using our framework presented previously, $APC$ and $APPC$ methods can be expressed as follows:
\begin{itemize}
    \item The solver $s$ being trained by $APC$ methods is an algorithm portfolio,
    while the solver $s$ being trained by $APPC$ methods is a parallel algorithm portfolio.
    \item The training instance settings for $APC$ methods and $APPC$ methods are the same as $AAC$ methods.

    \item Basically, $APC$ methods and $APPC$ methods
    seek to find algorithms that can cooperate with each other to form the portfolio, which  implies these algorithms need to
perform differently from each other.
The representative method of $APC$ methods, Hydra \cite{xu2010hydra, xu2011hydra}, adopts a simple and greedy strategy to achieve the cooperation between the algorithms.
Based on a parameterized algorithm, Hydra repeatedly finds an algorithm configuration
(or multiple configurations in \cite{xu2011hydra}) that
complements the current portfolio
to the greatest extent to add to the portfolio.
Another representative method, ISAC \cite{kadioglu2010isac,malitsky2012instance},
implements the cooperation
between the algorithms by explicitly
clustering the training instances into different parts based on normalized instance features, and then
assigning these clusters to different algorithm configurations (based on a parameterized algorithm).
In general, any idea that promotes the difference between the behaviours of algorithms can be helpful in constructing portfolios.
For instance, the idea of Negatively Correlated Search (NCS), which was proposed in our previous work [42], can be extended for constructing portfolios. NCS comprises multiple search processes (e.g., algorithms) that are run in parallel, and information is shared to explicitly encourage each search process to emphasize the regions that are not covered by others. In case of portfolio construction, NCS can be used to simultaneously find multiple algorithms whose coverages on the instances do not overlap each other. This construction (different from Hydra) is a one-step process, and (different from ISAC) it does not rely on the features of instances.


Both Hydra and ISAC call an $AAC$ method (to find the algorithm to add to the portfolio) and a $PAS$ method (to build an algorithm selector) as subroutines. Thus the experience collection and exploitation in them are done by their respective subroutines.

    The representative $APPC$ method ParHydra \cite{lindauer2016automatic}
    is similar to Hydra except that ParHydra has no $PAS$ method involved.
    Another $APPC$ method EPM-PAP \cite{tang2014population}
select algorithms from a pool of candidates to form a parallel portfolio of which the constituent algorithms would interact with each other at run time.
    Since directly evaluating all possible portfolios is computationally prohibitive, EPM-PAP adopts a specially designed metric to evaluate a portfolio,
based only on the individual performance of each constituent algorithm of the portfolio. In the training the experience $E$ produced by EPM-PAP is the performances of all candidate algorithms on the training instances, and the exploitation of $E$ is using enumeration equipped with the metric to find the best portfolio.

\end{itemize}




\subsection{Transfer Methods} 
\label{sub:transfer}
The last class of methods are transfer methods,
which explicitly extract useful
information from
solving processes of the training instances,
and use it to improve the performance of the solver
on the target instances.
The biggest difference between transfer methods
and the methods reviewed above is that the former
collect experience based on every single instance, while
the latter collect based on a set of instances.
Recall that $AAC$ methods run the candidate algorithms
on a set of instances to evaluate them, and $PAS$ methods
run the member algorithms on all training instances
to collect the training data.
On comparison, transfer methods
extract information from each individual training instance,
and the extraction of different instances is independent of each other.
Two specific transfer methods, dubbed XSTAGE \cite{BoyanM00} and
MEMETrans \cite{feng2015memetic}, are reviewed here
\footnote{ In the paper \cite{BoyanM00} that proposed XSTAGE, the term
XSTAGE is used to denote the composition of the training method
and the solver. We use XSTAGE here only to denote the training method.
The term MEMETrans is created by us to denote the
transfer method proposed by \cite{feng2015memetic} since the paper
does not give a term for this method.}.
Both of them enhance search-based solvers by
introducing high-quality solutions during the solving processes.
Using our framework presented previously, they can be expressed as:

\begin{itemize}
    \item The solver $s$ being trained by XSTAGE is a multi-restart
         local search algorithm in which
         a value function is used to determine the starting point
         for each local search process.
         This value function is actually a regression model, and its  coefficients are constantly refined through the
         whole algorithm run.


         The solvers $s$ being trained by MEMETrans are
         search-based meta-heuristics.

        \item XSTAGE collects training instances and
        target instances from the same benchmark, implicitly assuming
        these instances are similar enough to make the transfer useful.
        MEMETrans collects training instances and target instances
        from multiple heterogeneous benchmarks, and uses an instance
        similarity measure to determine given a target instance,
        which training instances will be used.

    \item Both XSTAGE and MEMETrans first run the solvers to solve the training instances. The experience $E$ produced by XSTAGE is the final value functions of these runs, and the experience $E$ produced by MEMETrans is the obtained solutions.

    To exploit $E$, XSTAGE combines all the value functions in a majority voting manner into one value function which will be used in the output solver, and MEMETrans learns a mapping from each solved training instance to its corresponding solution, and then combines all the mappings to be an initialization module that helps generate high-quality initial solutions for the output solver.

\end{itemize}

\section{LiangYi}
\label{LiangYisec}
The two main issues of offline training are selecting
the training instances and training the solvers.
The two-stage strategy,
which is widely adopted by existing training methods
such as $AAC$ methods \cite{adenso2006fine,hutter2009paramils,ansotegui2009gender,hutter2011sequential,lopez2011irace},
$PAS$ methods \cite{xu2008satzilla,gebser2011portfolio, leyton2009empirical, hutter2014algorithm, hoos2014claspfolio, lindauer2015autofolio, xu2012satzilla2012, kadioglu2011algorithm,malitsky2013boosting, malitsky2013algorithm, hoos2014claspfolio, lindauer2015autofolio, kotthoff2012evaluation},
$APC$ methods \cite{xu2010hydra, xu2011hydra, kadioglu2010isac, malitsky2012instance, lindauer2016automatic}
and transfer methods \cite{BoyanM00, feng2015memetic},
treat them in two independent phases.
However, intrinsically, these two issues are correlated.
From the perspective of developing solvers,
the greatest chance of the solver getting improved is on
those problem instances which the current solver cannot solve well
\footnote{This is also the key idea behind Hydra \cite{xu2010hydra, xu2011hydra},
in which the training always focuses on those hard instances
to the current solver. However, such instance importance adaptation
is still within a fixed training set.}.
Thus those hard instances for the solver are best suited as
the training instances.
However, during the training process,
the solver is being adapted to the training instances;
As the training proceeds, instances that are
previously appropriate for the solver may not be appropriate later on.
Using fixed training instances (as two-stage strategy methods do)
is actually not very helpful for improving the solver, especially after the solver has been adapted to the training instances,
and may result in the waste of computing resources.
A better strategy is to dynamically change the training instances
during the training process to keep them always appropriate (hard)
for the solver, so that the solver can be
continuously improved.

Currently, the training instances for two-stage methods are
directly selected from some benchmarks \cite{BoyanM00,feng2015memetic}, or are randomly
generated through some instance generators
\cite{hutter2009paramils,ansotegui2009gender,hutter2011sequential,xu2008satzilla,lindauer2015autofolio,xu2012satzilla2012,malitsky2013boosting,xu2010hydra,kadioglu2010isac,lindauer2016automatic},
based on an assumption
that the benchmarks and generators could represent the target scenarios
to which the solver is expected to be applied.
However, such assumption is not always true.
The commonly studied
benchmark instances and randomly generated instances
may lack diversity, be too simple, and rarely resemble real-world instances \cite{hooker1995testing,smith2015generating}.
Such risks could be avoided by dynamically changing the training instances.
First, this strategy selects training instances that
are never easy for the solver. Second, this strategy keeps changing the
training instances, which naturally introduces the diversity.

Based on the above considerations,
we propose a new training method, dubbed LiangYi
\footnote{The name "LiangYi" comes from the Taoism of Chinese philosophy.
Generally, it means two opposite elements of the world
that interact and co-evolve with each other.}.
Basically, LiangYi is a competitive co-evolutionary \cite{hillis1990co}
framework that alternately trains the solver and
searches for new training instances. It maintains a
set of algorithms and a set of instances, and each set strives
to improve itself against the other during the
evolutionary process.
The details of this framework are elaborated below.
\subsection{General Framework}
\label{generalframework}

For the sake of brevity,
henceforth we will use notations for the frequently used terms.
All the used notations are summarized in Table~\ref{table-1}.

\begin{table}
\footnotesize
\caption{Summary of the main notations }
\label{table-1}
\begin{tabular}{p{4cm} p{8.5cm}}
\hline
	$\mathcal{A}$ & Parameterized algorithm used by LiangYi \\
	$\Theta$ & Parameter configuration space of $\mathcal{A}$ \\
	$\Gamma$ & Target instance space \\
    $AP$   &   Algorithm population \\
    $IP$   &   Instance population  \\
    $AP_{k}$ & $AP$ at the initial stage of the $k$-th cycle of LiangYi \\
    $IP_{k}$ & $IP$ at the initial stage of the $k$-th cycle of LiangYi \\
    $N_{AP}$ & Number of algorithms in $AP$ \\
    $N_{IP}$ & Number of instances in $IP$\\
    $alg$  &   An algorithm belonging to $AP$ \\
    $ins$  &   An instance belonging to $IP$ \\
    $P(\mathrm{solver, instance\ set})$ &  Performance of the solver on the instance set\\
    $Aggr()$ & Aggregate function\\
    $C(AP,IP,alg)$ &  Contribution of $alg$ to the performance of $AP$ on $IP$\\
    $f_{AP}(alg)$ & Fitness of $alg$\\
    $f_{IP}(ins)$ & Fitness of $ins$\\

\hline
\end{tabular}
\end{table}

The form of the solver being trained by
LiangYi is a portfolio that runs all candidate
algorithms in parallel when solving an instance
\footnote{
Employing algorithms in parallel
to problem instances is an
emerging area in training solvers \cite{tang2014population,lindauer2016automatic}.
Note that running all algorithms in parallel is different from an
algorithm portfolio \cite{smith2009cross,kotthoff2014algorithm, tang2014population}, which typically involves some mechanism (e.g., selection \cite{smith2009cross,kotthoff2014algorithm}) to allocate computational resource to different algorithms.
}.
In the co-evolutionary framework,
the parallel portfolio is called an algorithm population ($AP$).
The reason for choosing an $AP$ rather
than a single algorithm as the solver is simple:
It is often the case \cite{xu2008satzilla, xu2011hydra, kadioglu2010isac, lindauer2015autofolio,tang2014population, peng2010population} that for a problem domain there is no single best algorithm for all possible instances. Instead, different algorithms perform well on different problem instances. Thus an $AP$ containing multiple complementary algorithms has the potential to achieve better overall performance than a single algorithm.
Similar with $APC$ methods (see Section
\ref{sub:automatic_portfolio_construction})
,
LiangYi builds the solver ($AP$) based on a parameterized algorithm.
Let $\mathcal{A}$ and $\Theta$ denote the parameterized algorithm used by LiangYi and the corresponding parameter configuration space of $\mathcal{A}$, respectively.
Any algorithm $alg$ in $AP$ satisfies that $alg \in \Theta$.
Let $\Gamma$ denote the instance space containing all the target instances of the
solver trained by LiangYi.
The training instance set maintained by LiangYi is called an instance population ($IP$),
and any instance $ins$ in $IP$ satisfies that $ins \in \Gamma$.

LiangYi adopts an alternating strategy to evolve $AP$ and $IP$.
More specifically, LiangYi first fixes $IP$ while it uses a training module to evolve $AP$ for some generations, and then it fixes $AP$ and uses an instance searching module to evolve $IP$ for some generations. This process is called a cycle of LiangYi and it will be repeated until some stopping criterion is met.
Let $AP_{k}$ and $IP_{k}$ denote the $AP$ and the $IP$ at the initial stage of the $k$-th cycle of LiangYi.
The pseudo code of LiangYi is outlined in
Algorithm Framework \ref{LiangYi}.
LiangYi first randomly initializes $AP$
as $AP_{1}$
and $IP$ as $IP_{1}$ (Lines 1-2).
During the $k$-th cycle,
LiangYi first evolves $AP$
from $AP_{k}$ to $AP_{k+1}$,
and then evolves $IP$
from $IP_{k}$ to $IP_{k+1}$.
After the $k$-th cycle,
LiangYi enters the $(k+1)$-th cycle
with $AP$ as $AP_{k+1}$
and $IP$ as $IP_{k+1}$ (Lines 4-8).
Finally,
when LiangYi is terminated,
the current $AP$ is
returned as the output solver (Line 9).

\renewcommand{\algorithmicrequire}{\textbf{Input:}}
\renewcommand{\algorithmicensure}{\textbf{Output:}}

\captionsetup[algorithm]{labelsep=colon}
\floatname{algorithm}{Algorithm Framework}
\begin{algorithm}
\small
\caption{LiangYi($N_{AP}, N_{IP}, params_{AP}, params_{IP}$)}
\label{LiangYi}
\algsetup{
linenosize=\small,
linenodelimiter=.
}
\begin{algorithmic}[1]
\REQUIRE Number of algorithms in $AP$, $N_{AP}$, number of instances in $IP$,
$N_{IP}$, set of the parameters which control $EvolveAlg$, $params_{AP}$, set of the parameters which control $EvolveIns$, $params_{IP}$
\ENSURE current $AP$ when LiangYi is terminated
\STATE $AP_{1} \leftarrow $  Randomly generate $N_{AP}$ algorithms $\in{\Theta}$
\STATE $IP_{1} \leftarrow $  Randomly generate $N_{IP}$ instances $\in{\Gamma}$
\STATE $k \leftarrow 1$
\WHILE{\textbf{not} $LiangYiTermination()$}
\STATE $AP_{k+1} \leftarrow EvolveAlg(AP_{k}, IP_{k}, params_{AP})$
\STATE $IP_{k+1} \leftarrow EvolveIns(AP_{k+1}, IP_{k}, params_{IP})$
\STATE $k \leftarrow k + 1$
\ENDWHILE
\RETURN{$AP_{k}$}
\end{algorithmic}
\end{algorithm}

\begin{figure}
\centering
\includegraphics[scale = 0.9]{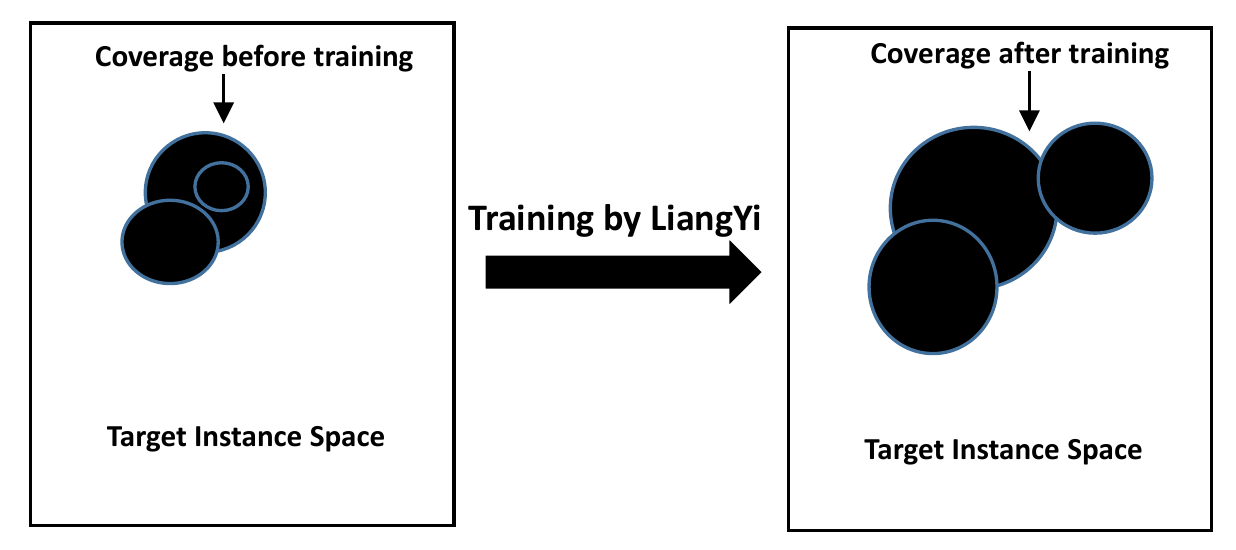}
\caption{An intuitive example of LiangYi enlarging the coverage of the
algorithm population on the target instance space.
The whole target instance space is the area inside the rectangle.
The $AP$ contains 3 member algorithms.
The coverage of each algorithm on the target instance space is indicated by a black circle.}
\label{figure-1}
\end{figure}

At each cycle of LiangYi,
the evolution of $AP$ is to improve the performance of $AP$ on $IP$ while keeping the good performances obtained by $AP$ in previous cycles, and the evolution of $IP$ is to discover those instances that cannot be solved well by $AP$ currently. Intuitively, if we consider an instance
in $\Gamma$ (the target instance space)
"covered" by a solver as it can be
solved well by the solver,
the essence of LiangYi is to enlarge the solver's
coverage on the target instance space
by
\begin{enumerate*}[label={\alph*)}]
    \item making the solver cover the area that has not been covered yet and
    \item keeping the solver covering the area that has already been covered.
\end{enumerate*}
Figure~\ref{figure-1} gives an intuitive visual example in which the $AP$ manages to cover a much larger area through training.


\subsection{Implementation Details}
\label{sec:implementation}

The training module and the instance searching module in LiangYi
are implemented as
two evolutionary procedures $EvolveAlg$ and $EvolveIns$,
respectively.
In general, any search method can be used in these two procedures.
In our work, evolutionary algorithms (EAs) \cite{back1996evolutionary} are employed as the
off-the-shelf tools, because EAs are suitable for handling populations (of either algorithms or instances)
and are less restricted by the properties of objective functions
(in comparison to other search methods such as gradient descent
that requires differentiable objective functions).
The behaviors of $EvolveAlg$
and $EvolveIns$ are controlled by parameter
sets $params_{AP}$ and $params_{IP}$, respectively.

When applying EAs to evolve $AP$ and $IP$,
the basic aspects that need to be considered are the representation, the variation (search operators such as crossover and mutation), the evaluation and the selection of the individuals in both populations.
Generally, the representation and the variation of the individuals greatly determine the search space for EA. In our work the expected search spaces for EAs are the parameter configuration space $\Theta$ and the target instance space $\Gamma$, which are actually strongly correlated to the used parameterized algorithm and the target problem class, respectively.
Thus in the framework of LiangYi, we do not
specify the individual representation and variation. When applying LiangYi,
the design issues in these two aspects should be addressed according to the target
problem class and the chosen parameterized algorithm.


Before going into the details of $EvolveAlg$ and $EvolveIns$, we explain here how to measure the performance of a population-based solver, which is the basis of the individual evaluation in both populations.
Let $P(\mathrm{solver, instance\ set})$
denote the performance of a solver on an instance set
according to a given performance metric $m$,
in which the solver could be a single algorithm or
an $AP$
and the instance set could be a single instance or an $IP$.
Since a population-based solver runs all member algorithms in
parallel when solving an instance,
the performance of $AP$
on an instance $ins$
is the best performance
achieved by its member algorithms on $ins$
(we assume a larger value is better for $m$ without loss of generality), i.e,
\begin{equation}
\label{equation1}
  P(AP,ins) = \max_{alg \in AP} P(alg, ins).
\end{equation}
The performance of $AP$ on $IP$
is an aggregated value of the performance of $AP$
on each instance of $IP$, i.e.,
\begin{equation}
\label{equation2}
  P(AP, IP) = \underset{ins \in IP}{Aggr}(P(AP, ins))
\end{equation}
where $Aggr()$ is an aggregate function.
The performance metric $m$ and the aggregate function $Aggr()$ are user-specified.

\subsubsection{Evolution of the Algorithm Population}
The pseudo code of $EvolveAlg$ is shown in Procedure \ref{evolvealg}.
\captionsetup[algorithm]{labelsep=colon}
\floatname{algorithm}{Procedure}
\begin{algorithm}
\small
\caption{$EvolveAlg(AP_{k}, IP_{k}, params_{AP})$}
$Memory$ is a global cache storing
the final $AP$ and the final performance matrix $M$ of $EvolveAlg$
in each cycle of LiangYi
\label{evolvealg}
\algsetup{
linenosize=\small,
linenodelimiter=.
}
\begin{algorithmic}[1]
\REQUIRE Algorithm population $AP_{k}$, instance population $IP_{k}$,
parameter set $params_{AP}$ containing:
set of parameters which control $VariationAlg$, $params_{VarA}$,
and set of parameters which control $RemoveWorst$, $params_{R}$
\ENSURE  current AP when $EvolveAlg$ is terminated
\STATE $M \leftarrow $ Test $AP_{k}$ on $IP_{k}$
\WHILE{\textbf{not} $EvolveAlgTermination()$}
\STATE $a_{1}, a_{2} \leftarrow $ Randomly select two individuals from $AP_{k}$
\STATE $a_{new} \leftarrow $ $VariationAlg (a_{1}, a_{2}, params_{VarA})$
\STATE $M_{new} \leftarrow$ Test $a_{new}$ on $IP_{k}$
\STATE $AP^{\prime}_{k} \leftarrow  AP_{k} \cup \{a_{new}\}$
\STATE $M^{\prime} \leftarrow $ Vertically concatenate $M$ and $M_{new}$
\STATE $AP_{k}, M \leftarrow  RemoveWorst(AP_{k}^{\prime}, M^{\prime}, params_{R})$
\ENDWHILE
\STATE Add $AP_{k}$ and $M$ to $Memory$
\RETURN{$AP_{k}$}
\end{algorithmic}
\end{algorithm}
First of all,
$AP_{k}$ is tested on $IP_{k}$ and
the result is represented
by a $N_{AP} \times N_{IP}$ matrix $M$ ($N_{AP}$ and $N_{IP}$ are the number of the algorithms in $AP$ and
the number of the instances in $IP$, respectively),
in which each row
corresponds to an algorithm in $AP_{k}$ and each column corresponds
to an instance in $IP_{k}$, so each entry in $M$
is the performance of the corresponding algorithm
in $AP_{k}$ on the corresponding instance in $IP_{k}$ (Line 1).
Then at each generation, two individuals
are randomly selected from $AP_{k}$ and an offspring
is generated by variation (Lines 3-4).
The generated algorithm $a_{new}$ is then tested on $IP_{k}$
and the result is represented
by a $1 \times N_{IP}$ matrix $M_{new}$ (Line 5).
The last step is to decide the survivors of this generation (Lines 6-8).
Together with all the algorithms in $AP_{k}$, $a_{new}$
is put into a temporary algorithm population $AP^{\prime}$.
The corresponding performance matrix $M^{\prime}$
to $AP^{\prime}$, of which the size is $(N_{AP}+1) \times N_{IP}$, is
constructed by concatenating $M$ and $M_{new}$ vertically.
Then the procedure $RemoveWorst$ is run to
decide which algorithm in $AP^{\prime}$ will be removed.

Basically, $RemoveWorst$ first calculates the fitness of each algorithm
in $AP^{\prime}$ and then selects the algorithm with
the lowest fitness to be removed.
The core of $RemoveWorst$ is
its fitness evaluation.
The idea is that an algorithm will
be preferred only if it contributes to $AP$,
and the more it contributes, the more it is preferred.
The contribution of a member algorithm is actually the performance improvement it brings to the population, which can be calculated as the population's performance loss caused by removing the algorithm.
Formally, let $C(AP, IP, alg)$ denote the contribution of algorithm $alg$
to the performance of $AP$ on $IP$.
If $|AP| > 1$, which means $AP$ contains other algorithms
besides $alg$,
$C(AP, IP, alg)$ is calculated via Equation (\ref{equation3}):
\begin{equation}
\label{equation3}
  C(AP, IP, alg) = |P(AP, IP) - P(AP-\{alg\}, IP)|,
\end{equation}
where $P(AP, IP)$ and $P(AP-\{alg\}, IP)$ are
calculated via Equation~(\ref{equation2}).
If $|AP| = 1$, which means $AP$ only contains one algorithm (i.e., $alg$),
and then removing $alg$ from $AP$ will cause complete performance loss
on $IP$. In this case, $C(AP, IP, alg)$ is calculated via Equation (\ref{equation4}):
\begin{equation}
\label{equation4}
  C(AP, IP, alg) = \alpha|P(AP, IP)|,
\end{equation}
where the parameter $\alpha > 0$.

Based on Equation~(\ref{equation4}),
for each member algorithm $alg$ in
the temporary algorithm population $AP_{k}^{\prime}$,
its performance contribution on $IP_{k}$
is $C(AP_{k}^{\prime}, IP_{k}, alg)$.
A high contribution indicates that the
corresponding algorithm
should be reserved to next generation.
However, directly using $C(AP_{k}^{\prime}, IP_{k}, alg)$
as the fitness of $alg$ is not appropriate.
As aforementioned (see Section \ref{generalframework}),
the evolution of $AP_{k}$ should not only improve the
performance of $AP_{k}$ on $IP_{k}$, but also
keep the good performances obtained
in previous cycles (on $IP_{1},...,IP_{k-1}$).
Using $C(AP_{k}^{\prime}, IP_{k}, alg)$ to evaluate
$alg$ only considers the first target.
Hence the fitness of $alg$ is calculated based on two types
of contributions:
The first one is the current performance contribution,
i.e., $C(AP_{k}^{\prime}, IP_{k}, alg)$,
and
the second one is the historical performance contribution of $alg$
on $IP_{1},...,IP_{k-1}$ (if there exist).

To calculate the historical performance contribution,
the concept $\mathit{age}$ is introduced
to describe how long an algorithm has
been in $AP$.
Suppose an algorithm $alg$ in $AP_{k}^{\prime}$ was added to $AP$
in the $j$-th cycle of LiangYi (and now is in the $k$-th cycle),
the $age$ of $alg$
is $k-j$.
The performance of $alg$
on $IP_{j}$,...,$IP_{k}$ are known
because $alg$ has been tested
on them in corresponding cycles of LiangYi
\footnote{Note that the performance contributions
of $alg$ on $IP_{1}$,...,$IP_{j-1}$ are not
considered in this paper
because the performances of $alg$ on them
are unknown.
To obtain these performances,
we can store $IP_{1}$,...,$IP_{j-1}$
and test $alg$ on them.
However, this would make the computational cost
and the storage cost increase fast over time.}.
To calculate the historical performance contribution
of $alg$ on $IP_{r} ( j \leq r \leq k-1 )$,
the algorithms that are in $AP_{k}^{\prime}$ and satisfy
the condition $age \geq (k-r)$
are selected (our target algorithm $alg$ is also selected, since its $age$ is $k-j$, which satisfies the condition) to form a virtual algorithm population
$virtualAP_{r}$.
The condition $age \geq (k-r)$
indicates that these selected algorithms were added
to $AP$ during or before
the $r$-th cycle of LiangYi, so they have been tested on $IP_{r}$.
The performances of these algorithms on $IP_{r}$
are represented by a $|virtualAP_{r}| \times N_{IP}$ matrix $virtualM_{r}$.
If $|virtualAP_{r}| > 1$, which means $virtualAP_{r}$ contains other algorithms
besides $alg$,
the performance contribution of $alg$ on
$IP_{r}$ is calculated via Equation~(\ref{equation3}):
\begin{equation*}
  C(virtualAP_{r}, IP_{r}, alg) = |P(virtualAP_{r}, IP_{r}) - P(virtualAP_{r}-\{alg\}, IP_{r})|.
\end{equation*}
If $|virtualAP_{r}| = 1$ , which means $virtualAP_{r}$ only contains $alg$,
and thus removing $alg$ from $virtualAP_{r}$ will
cause complete loss of performance on $IP_{r}$.
In this case, $C(virtualAP_{r}, IP_{r}, alg)$ is calculated
via Equation~(\ref{equation4}):
\begin{equation*}
  C(virtualAP_{r}, IP_{r}, alg) = \alpha|P(virtualAP_{r}, IP_{r})|.
\end{equation*}

Now we have all the performance contributions of $alg$ on $IP_{j}$,...,$IP_{k}$.
The fitness of $alg$, denoted as $f_{AP}(alg)$,
is calculated via Equation~(\ref{equation5})
\begin{equation}
\label{equation5}
  f_{AP}(alg) = \frac{\beta\Sigma_{ j \leq r \leq k-1}{C(virtualAP_{r}, IP_{r}, alg)} + C(AP^{\prime}_{k}, IP_{k}, alg)} {k-j+1}
\end{equation}
where $k$ is the index of the current cycle of $LiangYi$, $j$ is the $age$ of $alg$, and $\beta$ is a nonnegative parameter.
The terms $C(virtualAP_{r}, IP_{r}, alg) (j \leq r \leq k-1)$ are historical
performance contributions on $IP_{j}$,...,$IP_{k-1}$,
while $C(AP^{\prime}_{k}, IP_{k}, alg)$ is
the current performance contribution on $IP_{k}$.
Thereby the numerator in the fraction is
actually a weighted sum of $(k-j+1)$ performance contributions, in which
the parameter $\beta$ is used to balance between historical performance contributions(on $IP_{1},...,IP_{k-1}$) and current performance contribution (on $IP_{k}$).

The pseudo code of $RemoveWorst$ is
demonstrated in Procedure \ref{removeworst}.
First the fitness of each algorithm in
$AP^{\prime}_{k}$ is calculated (Lines 1-15).
Specifically, for an algorithm $alg$ which was
added to $AP$ in the $j$-th cycle of LiangYi,
$k-j$ virtual algorithm populations, i.e.,
$virtualAP_{j},...,virtualAP_{k-1}$,
are constructed (according to the global cache Memory) to calculate its
historical performance
contributions on $IP_{j},...,IP_{k-1}$,
via Equation~(\ref{equation3}) or Equation~(\ref{equation4}) (Lines 2-12).
Together with the current performance
contribution calculated via Equation~(\ref{equation3}) (Line 13),
the historical contributions are used to calculate the fitness of
$alg$ via Equation~(\ref{equation5}) (Line 14).
After the fitness of each algorithm in $AP^{\prime}_{k}$
has been calculated,
the algorithm with the lowest fitness will be removed (Line 16).

\captionsetup[algorithm]{labelsep=colon}
\floatname{algorithm}{Procedure}
\begin{algorithm}
\small
\caption{$RemoveWorst(AP^{\prime}_{k}, M^{\prime}, params_{R})$}
\label{removeworst}
\algsetup{
linenosize=\small,
linenodelimiter=.
}
\begin{algorithmic}[1]
\REQUIRE Temporary algorithm population $AP^{\prime}_{k}$,
temporary performance matrix $M^{\prime}$, parameter set $params_{R}$ containing
$\alpha$ (used in Equation~(\ref{equation4}))
and
$\beta$ (used in Equation~(\ref{equation5}))
\ENSURE  $AP_{k}$, performance matrix $M$
\FOR{\textbf{each} algorithm $alg$ in $AP^{\prime}_{k}$}
\STATE $age \leftarrow$ Query $Memory$ for how many cycles $alg$ has been staying
\STATE $j \leftarrow k - age$
\FOR{$ r \leftarrow j$ to {$k-1$}}
\STATE $VirtualAP_{r}\leftarrow $ Select algorithms which satisfy the condition $age \geq (k-r) $ (according to $Memory$) from $AP^{\prime}_{k}$
\STATE $M_{r} \leftarrow $ Construct corresponding performance matrix
to $VirtualAP_{r}$ (according to $Memory$)
\IF{$|virtualAP_r| > 1$}
\STATE $C(virtualAP_r, IP_{r}, alg) \leftarrow $ Calculate
the algorithm contribution of $alg$ on $IP_{r}$ via Equation~(\ref{equation3})
\ELSE
\STATE $C(virtualAP_r, IP_{r}, alg) \leftarrow $ Calculate
the algorithm contribution of $alg$ on $IP_{r}$ via Equation~(\ref{equation4})
\ENDIF
\ENDFOR
\STATE $C(AP^{\prime}_{k}, IP_{k}, alg) \leftarrow $ Calculate
the algorithm contribution of $alg$ on $IP_{k}$ via Equation~(\ref{equation3})
\STATE $f_{AP}(alg) \leftarrow $ Calculate the fitness of $alg$ via
Equation~(\ref{equation5})
\ENDFOR
\STATE $AP_{k}, M \leftarrow $ Remove the algorithm with the
lowest fitness from $AP_{k}^{\prime}$ and the corresponding row from $M^{\prime}$
\RETURN{$AP_{k}$, $M$}
\end{algorithmic}
\end{algorithm}

\subsubsection{Evolution of the Instance Population}
As aforementioned (see Section \ref{generalframework})
the evolution of $IP$
aims at discovering those
instances that cannot be solved well by $AP$;
thus the fitness of an instance in $IP$ is measured by
how $AP$ performs on it ---
the worse the performance, the higher the fitness.

The pseudo code of $EvolveIns$ is demonstrated in
Procedure \ref{evolveins}.
First of all,
$AP_{k+1}$ is tested on the
$IP_{k}$ and the result is
represented by a $N_{AP} \times N_{IP}$ matrix $M$ (Line 1),
and the fitness of each instance is calculated (Lines 2-4).
The fitness of an instance $ins$, denoted as $f_{IP}(ins)$,
is calculated via Equation~(\ref{equation6}):
\begin{equation}
\label{equation6}
f_{IP}(ins) = -P(AP_{k+1}, ins),
\end{equation}
where $P(AP_{k+1}, ins)$ is the performance of $AP_{k+1}$ on instance $ins$,
calculated via Equation~(\ref{equation1}).
At each generation,
$N_{IP}*re$ new instances are generated by
repeatedly selecting two instances from $IP_{k}$ (using tournament selection \cite{back1996evolutionary}) and
creating two offsprings by variation (Lines 6-11).
These offsprings are then tested
against the algorithm population $AP_{k+1}$ and
the fitness of each offspring is calculated (Lines 12-15).
At the end of this generation,
all instances in $IP_{k}$ and the offsprings are put into a candidate pool
and the worst $N_{IP}*re$ instances are removed (Lines 16-19).
\captionsetup[algorithm]{labelsep=colon}
\floatname{algorithm}{Procedure}
\begin{algorithm}
\small
\caption{$EvolveIns(AP_{k+1}, IP_{k}, params_{IP})$}
\label{evolveins}
\algsetup{
linenosize=\small,
linenodelimiter=.
}
\begin{algorithmic}[1]
\REQUIRE Algorithm population $AP_{k+1}$, instance population $IP_{k}$,
parameter set $params_{IP}$ containing:
set of parameters which control $VariationIns$, $params_{VarI}$, and
replacement ratio $res$
\ENSURE current $IP$ when $EvolveIns$ is terminated
\STATE $M \leftarrow $ Test $AP_{k+1}$ on $IP_{k}$
\FOR{\textbf{each} instance $ins$ in $IP_{k}$}
\STATE $f_{IP}(ins) \leftarrow $ Calculate the fitness of $ins$
via Equation~(\ref{equation6})
\ENDFOR
\WHILE{\textbf{not} $EvolveInsTermination()$}
\STATE $offsprings \leftarrow \varnothing$
\FOR{$i \leftarrow 1$ to $\frac{N_{IP}*res}{2}$}
\STATE $ins_{1}, ins_{2} \leftarrow$ Select two parents from $IP_{k}$ with tournament selection
\STATE $ins_{new1}, ins_{new2} \leftarrow VariationIns(ins_{1}, ins_{2}, params_{VarI}) $
\STATE $offsprings \leftarrow offsprings \cup \{ins_{new1}, ins_{new2}\}$
\ENDFOR
\STATE $M_{offsprings} \leftarrow$ Test $AP_{k+1}$ on $offsprings$
\FOR{\textbf{each} instance $ins$ in $offsprings$}
\STATE $f_{IP}(ins) \leftarrow $ Calculate the fitness of $ins$
via Equation~(\ref{equation6})
\ENDFOR
\STATE $candidates \leftarrow IP_{k} \cup offsprings$
\STATE $M_{candidates} \leftarrow $ Horizontally concatenate $M$ and $M_{offspring}$
\STATE $IP_{k}, M \leftarrow$ Remove the worst $N_{IP} * re$ instances from $candidates$ and the corresponding columns from $M_{candidates}$
\ENDWHILE
\RETURN{$IP_{k}$}
\end{algorithmic}
\end{algorithm}

\section{Case Study: the Travelling Salesman Problem}
\label{sec:empiricalstudy}

The main purpose of this section is to empirically verify
whether LiangYi is an effective method for improving solvers.
We evaluated LiangYi
on the Travelling Salesman Problem (TSP) \cite{lawler1985traveling},
one of the most well-known computationally hard
optimization problem.
Specifically, the symmetric TSP, i.e., the distance between
two cities is the same in each opposite direction,
with Euclidean distances in a two-dimensional space
is considered here.
In the remainder of this section, we first
give the target scenario (including the
target instances and the performance metric) where LiangYi is applied,
and then instantiates LiangYi for the scenario.
After that, we first compare LiangYi to other existing training methods,
and then we investigate the properties of LiangYi to see whether it is able
to perform as expected.

All of our experiments
were carried out on a workstation of
two Xeon CPU with 24 cores and 48 threads at 2.50GHz,
running Ubuntu Linux 16.04.


\subsection{Target Scenario}
\label{targetscenario}
The target instances considered here
are all
TSP instances with problem size equal to 500,
i.e., the number of cities equals to 500.
This work focuses on optimizing
the solver's applicability on the target instances, i.e.,
the performance metric is applicability.
A solver is said to be applicable to
an instance if it can find a good enough solution
to this instance within a given time.
For TSP, the goodness of a solution $sol$ is measured by
the percentage by which the tour length of $sol$
exceeds the tour length of the optimum $sol^{\star}$
\footnote{The optimum $sol^{\star}$ is obtained using Concorde \cite{applegate2006concorde},
a branch-and-cut based exact TSP solver.}, abbreviated as PEO(percentage excess optimum):
\begin{equation*}
  PEO = \frac{lengh(sol) - lengh(sol^{\star})}{lengh(sol^{\star})} * 100\%.
\end{equation*}
With the definition of PEO, given a cut-off time $t$,
a solver is said to be \textit{applicable} to an instance $ins$
if the $PEO$ of the best solution found by the solver
in time $t$
is below a threshold $\theta$.
With the definition of the applicability of a solver
to a single instance, the applicability of a solver to an instance set is defined as the fraction of the instances to which the solver is applicable.

In this paper very radical values for the
cut-off time $t$ and the PEO threshold $\theta$ are
adopted ($t=0.1s$ and $\theta = 0.05\%$)
to see whether LiangYi is able to evolve solvers that can
work well under such harsh conditions.

\subsection{Instantiation of LiangYi and Its Computational Cost}

In order to instantiate LiangYi for the above scenario,
there are several issues to be addressed.
The first issue is to specify the performance function
$P(\mathrm{solver}, \mathrm{instance\ set})$ used by LiangYi (see Section \ref{sec:implementation}) so that LiangYi can optimize
the applicability appropriately.
The performance of an algorithm $alg$
on an instance $ins$, i.e., $P(alg,ins)$
in Equation~(\ref{equation1}),
is specified as follow:
\begin{equation*}
P(alg, ins)=
\begin{cases}
1,\ \ \mathrm{if}\ alg \ \mathrm{is}\ \mathrm{applicable}\ \mathrm{to}\ ins \\
0,\ \ otherwise.
\end{cases}
\end{equation*}
Intuitively, an $AP$ is said to be
applicable to an instance $ins$ if any algorithm
of $AP$ is applicable to $ins$.
With $P(alg,ins)$ specified as above,
this definition is equivalent to the definition
given by Equation~(\ref{equation1}), namely, $AP$
is applicable to $ins$ if the best algorithm of $AP$
is applicable to $ins$.
The aggregate function $aggr()$ in
Equation~(\ref{equation2}) is specified as
returning the mean value of the aggregated terms:
\begin{equation*}
  P(AP,IP) = \frac{\sum_{ins \in IP}P(AP,ins)}{|IP|},
\end{equation*}
which essentially calculates the
proportion of the instances to which $AP$ is applicable.

\begin{table}[t]
\footnotesize
\caption{The parameters of the CLK used}
\begin{center}
\begin{tabular}{p{4cm} p{2.5cm} c}
  \hline
  Parameters      &   Parameter Type  & \# of Candidate Values  \\
  \hline
  Initialization Strategy & Categorical & 4 \\
  Perturbation Strategy & Categorical   & 4\\
  Search Depth     &  Numerical  & 6\\
  Search Width       & Numerical & 8\\
  Backtrack Strategy  & Categorical & 14\\
 \hline
\end{tabular}
\end{center}
\label{table0}
\end{table}

The second issue is to choose a parameterized algorithm
for LiangYi to build an $AP$ based on it.
The choice of the parameterized algorithm in this work
is Chained Lin-Kernighan (CLK) \cite{applegate2003chained}.
It is a variant of the Lin-Kernighan heuristic \cite{lin1973effective},
one of the best heuristics for
solving symmetric TSP.
CLK
chains multiple runs of the Lin-Kernighan algorithm to
introduce more robustness in the resulting tour.
Each run starts with a perturbed version of the final tour of the
previous run.
We extended the original implementation of CLK
to allow a more comprehensive control of its components.
The parameters of the resulting algorithm are summarized
in Table~\ref{table0}.
To handle the randomness of CLK,
we adopt a simple way - fixing the random seed
of CLK and turning it into a deterministic algorithm.
To use CLK in the target scenario (see Section~\ref{targetscenario}), in our experiments we always set the runtime of CLK to 0.1s, and after it was terminated we checked whether PEO of the solution found was below 0.05\%.

The third issue is to
specify the representation and the variation of individuals in both populations (see Section \ref{sec:implementation}).
Each algorithm in $AP$
is represented by a list containing 5 integers, each of which
indicates its value for the corresponding parameter.
Each instance in $IP$ is represented
by a list of 500 $(x,y)$ coordinates on a $ 10^{6} \times 10^{6}$ grids.
The random initialization for $AP$ and $IP$ works
by uniformly randomly selecting a value
(i.e., a parameter value for the algorithm,
or two coordinates for the instance)
from candidate values
for each entry of the individual (the algorithm or the instance).
Both the variation for the individuals in $AP$ and the variation for
the individuals in $IP$ are implemented as a uniform crossover and a uniform mutation \cite{back1996evolutionary}.
The uniform crossover operates with a probability,
by choosing for each entry of the offspring with equal probability
either the value of the entry from the first or the second parent.
The probability of the uniform crossover being operated, i.e., crossover probability,
is controlled by parameters $cro_{alg}$ (in $AP$) and $cro_{ins}$ (in $IP$).
The mutation consists of
replacing the value of each entry of the offspring,
with a probability (mutation rate),
with uniformly randomly chosen one from the candidate values.
The mutation rate is controlled by parameters $mu_{alg}$ (in $AP$) and $mu_{ins}$ (in $IP$).


\begin{table}[t]
\footnotesize
\caption{The parameter settings of the instantiation of LiangYi for TSP}
\begin{center}
\begin{tabular}{p{3.2cm} p{2.1cm}}
  \hline
  {$EvolveAlg$}      &   {$EvolveIns$}     \\ \hline
  $N_{AP} = 6$       & $N_{IP} = 150$  \\
  \# of Gens = 500   &  \# of Gens = 10 \\
  $cro_{alg} = 0.6 $ & $cro_{ins} = 1  $   \\
  $mu_{alg} = 0.6 $  & $mu_{ins} =  0.8$   \\
  $\alpha = 2$       & $res = 0.3      $   \\
  $\beta =  2$       &                     \\
 \hline
\end{tabular}
\end{center}
\label{table1}
\end{table}

The last issue is to set the termination conditions and the parameters of LiangYi.
The termination condition for LiangYi is
the number of cycles reaching 3.
In each cycle, procedure
$EvolveAlg$ will be run for 500 generations and
$EvolveIns$ will be run for 10 generations.
The number of algorithms in $AP$, i.e., $N_{AP}$, is set to 6,
and the number of instances in $IP$, i.e., $N_{IP}$, is set to 150.
The parameter settings of LiangYi are listed in Table~\ref{table1}.
In order to maintain a high level of diversity within
$AP$,
the mutation rate in $evolvealg$ is set to a high value (0.4).
For $evolveins$, it is important to keep
the instance population exploring the target instance space
instead of stagnating in some local areas,
and therefore the mutation rate in $evolveins$ is
also set to a high value (0.7).

\begin{figure}
\centering
\includegraphics[scale = 1.0]{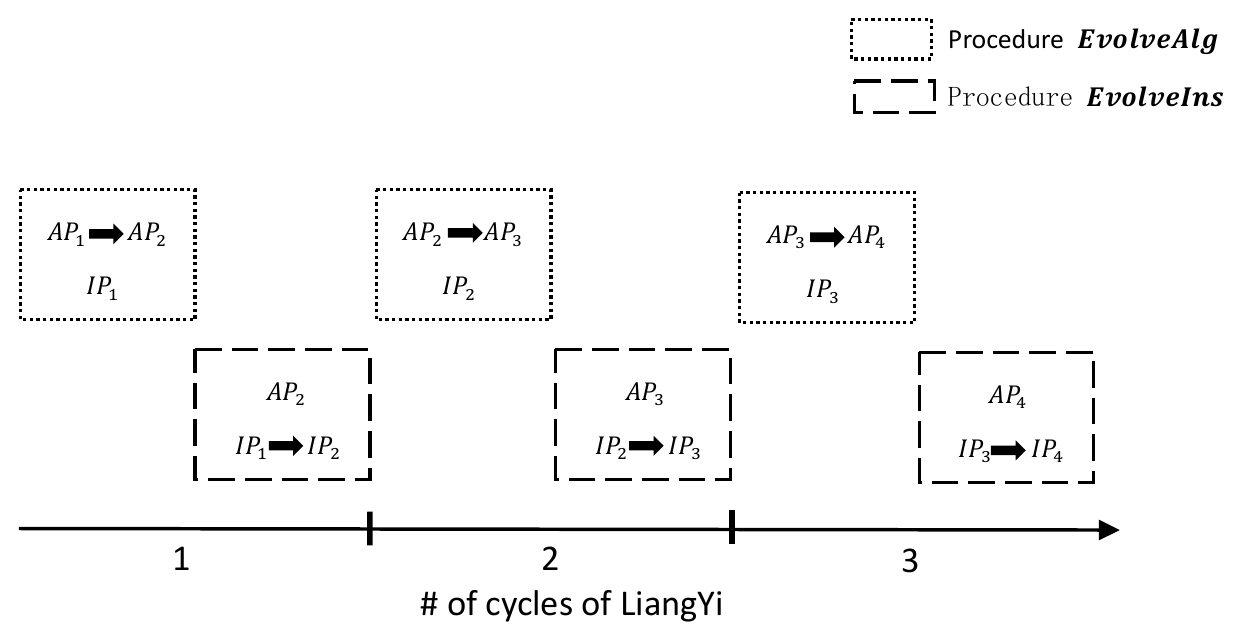}
\caption{The training process of LiangYi in which AP and IP are evolved alternatively.
Dotted rectangle and dashed rectangle represent procedure $EvolveAlg$ and
procedure $EvolveIns$, respectively.}
\label{figure0}
\end{figure}

We applied the instantiation of LiangYi described above
to the considered scenario. The training process in which $AP$ and $IP$ are evolved alternatively is depicted in Figure~\ref{figure0}.
The computational cost of LiangYi is mainly composed of two parts: The first part is the overhead for the algorithm runs used to evaluate algorithms (in $EvolveAlg$) or instances (in $EvolveIns$); The second part is the overhead for solving exactly the instances to obtain their optima (in $EvolveIns$).
For each cycle of $LiangYi$, in $EvolveAlg$ there are $(N_{AP}+AP_{G})* N_{IP}$ algorithm runs, and in $EvolveIns$ there are $(N_{IP}+IP_{G}*N_{IP} *res)*N_{AP}$ algorithm runs ($AP_{G}$ and $IP_{G}$ are the number of generations of $EvolveAlg$ and $EvolveIns$, respectively)
and meanwhile there are $(N_{IP}+IP_{G}*N_{IP} *res)$ instances to be solved exactly.
In our experiments, the time for each algorithm run was set to 0.1s (see Section~\ref{targetscenario}),
and the average time for exactly solving an instance
was 32 seconds (specific time varied from 1 second to 15 minutes).
Thus the estimated CPU time for one run of the instantiation of LiangYi was 81450 seconds (i.e., 22.6 hours).

\subsection{Comparative Study}
\label{comparative}
In this section we compare LiangYi with other existing training methods
in the considered scenario.
Since the $AP$ built by LiangYi
is actually a parallel portfolio,
we chose ParHydra \cite{lindauer2016automatic}
(see Section \ref{sub:automatic_portfolio_construction})
, the state-of-the-art automatic parallel portfolio
construction method, as the method to compare with.

\subsubsection{Settings of ParHydra}

ParHydra accepts a parameterized algorithm,
a set of training instances,
and a performance metric to be optimized.
For the target scenario considered,
the performance metric is applicability.
The parameterized algorithm fed to ParHydra
is CLK, same as LiangYi.
We used two different ways to construct the training sets
for ParHydra.
The first training set $IP_{random}$ was built according to
the usual practice for two-stage methods --- randomly generating
a set of instances.
Specifically, each instance in $IP_{random}$ was
generated by randomly choosing two
coordinates for each city on a $10^{6} \times 10^{6}$ grids.
The second training set $IP_{training}$ was built by collecting
the instance populations which were produced by LiangYi and once
served as the training instances during the training process, i.e.,
$IP_{training} = IP_{1} \cup IP_{2} \cup IP_{3}$.
Since LiangYi produces instance populations as by-products,
it is interesting to see
how good these instance populations are
as training instances for existing methods
like ParHydra. Both sets contain $450$ instances.
The solvers output by ParHydra based on $IP_{random}$ and $IP_{training}$ are denoted
as $PH_{random}$ and $PH_{LiangYi}$, respectively.

ParHydra is an iterative method, which builds the portfolio
from scratch and
adds an algorithm to it in each
iteration (See Section~\ref{sub:automatic_portfolio_construction}).
Thus we set the iteration number of ParHydra to 6
to keep in line with LiangYi in terms of algorithm
number. ParHydra iteratively calls an algorithm configurator as a subroutine. In our experiments
we used ParamILS \cite{hutter2009paramils}
(version 2.3.8 with its default instantiation
of FocusedILS with adaptive capping).
Since the implementation of ParamILS does not provide an option to directly optimize applicability,
we set the metric used in ParamILS to penalized average runtime, PAR1000
\footnote{ PAR1000 penalizes each unsuccessful run
(meaning not satisfying the PEO constraint in our experiments, see Section~\ref{targetscenario})
with 1000 times the given cut-off time (0.1s).
For each successful run, the run time is the given cut-off time (0.1s).},
in our setting which is equivalent to optimizing applicability.
At each iteration of ParHydra,
15 copies of ParamILS \cite{hutter2009paramils}
were run with different random seeds
in parallel to obtain 15 candidate algorithms, and
the one achieving the best performance on the training set was added to the portfolio.
The termination condition for each ParamILS run
was the total run time for configured algorithm (CLK)
reaching 5 hours. Thus the estimated total computation cost for
a run of ParHydra was 450 CPU hours (32.14 days).




\subsubsection{Experimental Protocol}
Since LiangYi and ParHydra are both stochastic methods,
we ran each comparative method 20 times and compare their test results.
Specifically, first we ran LiangYi 20 times, and therefore we obtained
20 $IP_{training}$ from these runs.
Then we randomly generated 20 different $IP_{random}$,
and based on each of these 40 training set (including $IP_{random}$ and $IP_{training}$),
we ran ParHydra to obtain a solver.
The random seeds used in the training processes of $PH_{random}$
are the same as the ones used in the training processes of $PH_{random}$.
Finally in total we obtained 20 $AP_{4}$ (the $AP$ output by LiangYi),
20 $PH_{random}$ and 20 $PH_{LiangYi}$.
In order to adequately assess the performances of these output solvers
in the target scenario,
we generated a huge test set, denoted as $IP_{test}$, containing
10000 TSP instances with the number of cities equal to 500.
Specifically,
each instance was generated by randomly choosing two coordinates for each city
from the interval $[0, 10^{6})$.
To our knowledge, this is the first time that a test set
of such a large size (10000) is used to test TSP solvers.

The runtime requirements in CPU days were as follows:
18.9 days for LiangYi  training (including 20 runs);
642.8 days for ParHydra training (including 20 runs);
7.80 days for testing (including the time for obtaining the
optima of the test instances).

\subsubsection{Experiment Results}
The average test results of these three types of solvers
(with each type containing 20 solvers)
are presented in Table~\ref{table3}.
Since for each $AP_{4}$ there is a $PH_{LiangYi}$
that shares the training instances with it, we
performed a two-sided Wilcoxon signed-rank test
to check whether the difference between the results obtained by
$AP_{4}$ and $PH_{LiangYi}$ are statistically significant.
We also performed a two-sided Wilcoxon signed-rank test
for $PH_{random}$ and $PH_{LiangYi}$, since they share the common
random seeds.
For $AP_{4}$ and $PH_{random}$, we performed a two-sided Wilcoxon rank-sum test.
All the tests were carried out with a 0.05 significance
level.
The statistical test results are also presented in Table~\ref{table3}.
$PH_{LiangYi}$ obtained better results than $PH_{random}$,
indicating the training instances produced by LiangYi
are more representative of the target scenario
than the randomly generated ones.
It is a little surprising to see $AP_{4}$ obtained
better results than $PH_{LiangYi}$ at the first sight.
Different from $AP_{4}$,
$PH_{LiangYi}$ was directly trained with the whole $IP_{training}$,
which was produced by LiangYi cycle by cycle;
thus it is conceivable that $PH_{LiangYi}$ would obtain
better results on $IP_{training}$ than $AP_{4}$
(actually their performances on $IP_{training}$, i.e.,
$P(AP_{4}, IP_{training})$ and $P(PH_{LiangYi}, IP_{training})$,
are 0.6063 and 0.6644).
The reason why $AP_{4}$ obtained better results than $PH_{LiangYi}$
on $IP_{test}$ is as follow: The adaptive instance updating strategy
used by LiangYi can be seen as a filter that
only reserves those hard instances for $AP$ to make the training
focus on them, which makes the actual coverage of $AP$
on the target instances far greater than
its coverage on $IP_{training}$ ($0.7001 > 0.6063$),
because those easy target instances which are
sampled by $EvolveIns$ and are actually covered by $AP$
are all filtered out.
Compared to LiangYi, ParHydra
accepts all the training instances and
only focuses on
the training set.
The lack of the instance adaptability
makes the performance of the output solver greatly
depend on how much the training set can represent the
target scenario.

In Table~\ref{table4}, we also give the average
PEO (see Section~\ref{targetscenario}) obtained by the three types of solvers
on $IP_{test}$.
$AP_{4}$ is still significantly better than
$PH_{LiangYi}$ and $PH_{random}$.
Although in the target scenario we actually
did not directly optimize the solution quality,
LiangYi managed to evolve solvers that on average
satisfy the PEO
requirements in the scenario (0.05\%).


\begin{table}[htbp]
  \footnotesize
  \centering
  \caption{Test result comparisons of $AP_{4}, PH_{random}$ and $PH_{LiangYi}$.
The test results are presented in terms of applicability,
i.e., the proportion of the
instances to which the solver is applicable.
We performed statistical tests to compare their results,
considering p-values below 0.05 to be statistically significant.
The last two columns provide the results of the test,
where $'W'$, $'D'$, and $'L'$ indicate the corresponding solver
is superior to, not significantly different from or inferior to the
competitor, respectively.}
    \begin{tabular}{l|c|c|c}
    \hline
          & \multirow{2}[0]{*}{Average test results (applicability)} &  \multicolumn{2}{c}{Statistical test results} \\
          &       & vs.$PH_{LiangYi}$ & vs.$PH_{random}$\\
    \hline
    $AP_{4}$    		& $0.6822 \pm 0.0082$ & W     & W \\
    $PH_{LiangYi}$    	& $0.6383 \pm 0.0138$ & $\pmb{-}$   & W \\
    $PH_{random}$    	& $0.6290 \pm 0.0133$ & L     & $\pmb{-}$  \\
    \hline
    \end{tabular}%
  \label{table3}%
\end{table}%

\begin{table}[htbp]
  \footnotesize
  \centering
  \caption{Average PEO (see Section \ref{targetscenario}) comparisons of $AP_{4}, PH_{random}, PH_{LiangYi}$ on $IP_{test}$. We performed  statistical tests here similar to the ones in the comparisons of average test results (applicability).}
    \begin{tabular}{l|c|c|c}
    \hline
          & \multirow{2}[0]{*}{Average PEO} &  \multicolumn{2}{c}{Statistical test results} \\
          &       & vs.$PH_{LiangYi}$ & vs.$PH_{random}$\\
    \hline
    $AP_{4}$    		& $0.0497\%   \pm 0.0010\%   $   &     W     	& W \\
    $PH_{LiangYi}$    	& $0.0532\%   \pm 0.0018\%   $   & 	$\pmb{-}$   & D \\
    $PH_{random}$    	& $0.0539\%   \pm 0.0015\%   $   &     D     	& $\pmb{-}$  \\
    \hline
    \end{tabular}%
  \label{table4}%
\end{table}%

\subsection{Investigating the Properties of LiangYi}
\label{verifying}
As aforementioned,
the idea behind LiangYi is
to optimize the performance of $AP$
on target instances by
a) improving its performance on those instances on which it performs badly and
b) keeping its good performance on those instances on which it performs well.
The main purpose of this section is to investigate
whether LiangYi is able to accomplish the two
objectives listed above.
Specifically, the verification is divided into two parts --- the training part and the test part.
In the training part
we investigate that, in the training process, whether LiangYi gives satisfactory answers to the following three questions:
\begin{enumerate}
    \item Is procedure $evolvealg$ able to improve the performance of $AP$ on current $IP$?
    \item Is procedure $evolveins$ able to degrade the performance of $AP$ on current $IP$?
    \item Is procedure $evolvealg$ able to keep the
    performance of $AP$ on previous $IPs$?
\end{enumerate}
The second question indicates whether the evolution of $IP$
is able to discover and include hard-to-solve instances
to $AP$, and the first question indicates
whether the evolution of $AP$ is able to
improve the performance of $AP$ on the hard instances included in the current $IP$.
The combination of these two checks the
whether LiangYi is able to accomplish the first objective.
The third question checks whether LiangYi is able to accomplish the second objective.
In addition to focusing on the three specific aspects,
we also directly check if LiangYi is able to continuously
improve $AP$ in the training part.
Specifically, we check whether the performance of $AP$
on $IP_{training}$ that are produced in the training are improved by LiangYi.
Similarly, in the test part we also directly
check whether the performance of $AP$ at the optimization task
is being improved by LiangYi.


\subsubsection{Training Part}
\label{training}

\begin{figure}
\centering
\includegraphics[scale = 0.6]{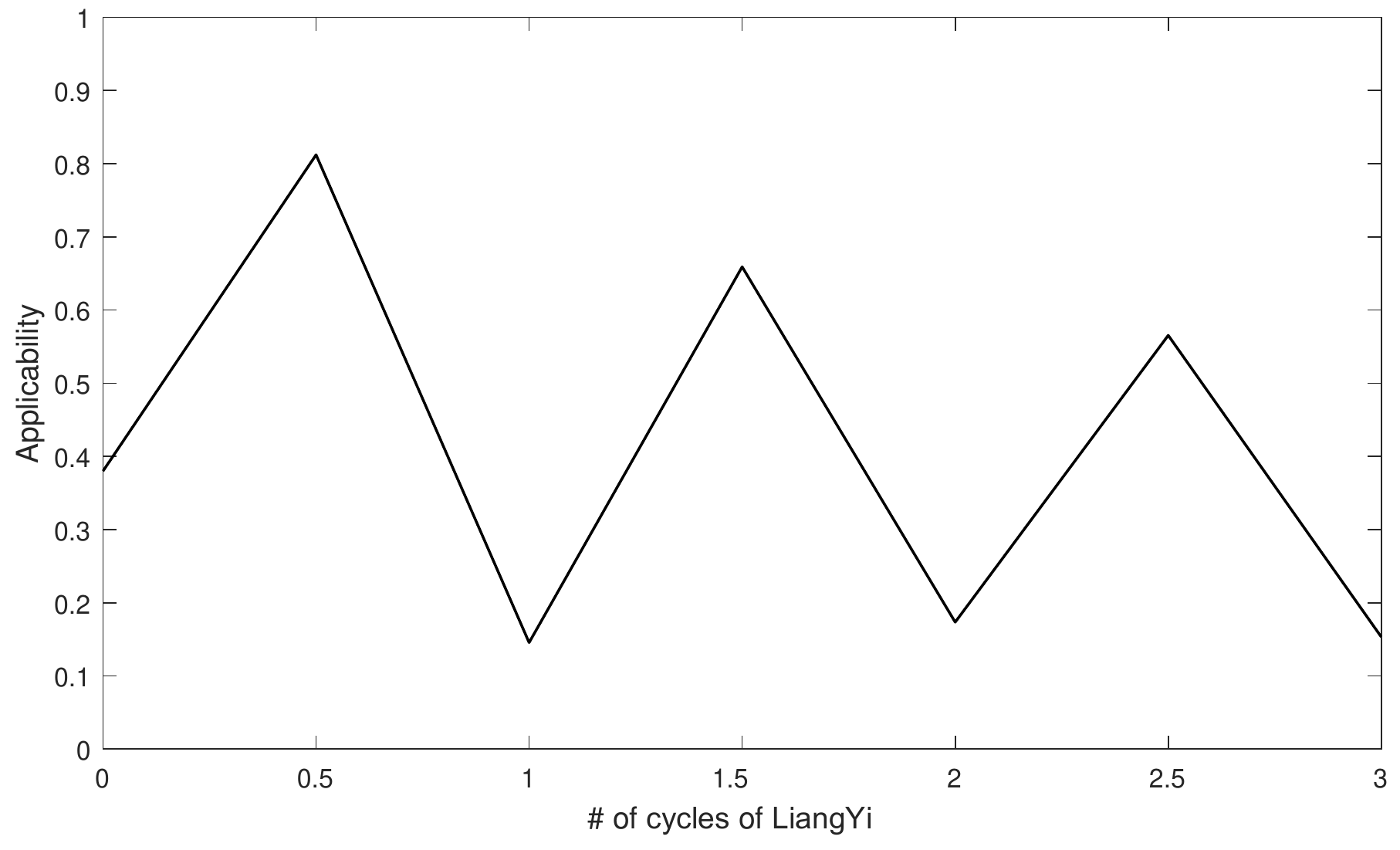}
\caption{The performance of $AP$ on $IP$
at the beginning ($EvolveAlg$ begins), the
middle ($EvolveAlg$ ends)
and the end ($EvolveIns$ ends) of each cycle of LiangYi.
The performances are represented in terms
of applicability, i.e., the proportion of the instances to which AP is
applicable, and are averaged over 20 runs.}
\label{figure1}
\end{figure}

To answer the first question and the second question,
the performance of
$AP$ on $IP$
during the training process
averaged over 20 runs are plotted in Figure~\ref{figure1}.
The results depicted in Figure~\ref{figure1}
clearly show that, at each cycle of LiangYi,
$EvolveAlg$ improves the performance of $AP_{k}$ on $IP_{k}$,
and $evolveins$ degrades the performance of $AP_{k+1}$ on $IP_{k}$,
which gives positive answers to the first two questions, thus
confirming the first aspect of the idea behind LiangYi.

The third question is answered in this way:
Since procedure $evolvealg$ evolved $AP_{k}$ to $AP_{k+1}$ to improve
the performance on $IP_{k}$, we checked whether the
improvement from
$P(AP_{k}, IP_{k})$ to $P(AP_{k+1}, IP_{k})$, i.e.,
$|P(AP_{k},IP_{k})-P(AP_{k+1},IP_{k})|$,
was kept
in subsequent cycles of LiangYi.
Specifically,
we tested $AP_{k+2},...,AP_{4}$
on $IP_{k}$ to obtain their performances
on $IP_{k}$, i.e., $P(AP_{k+2}, IP_{k}),...,P(A_{4}, IP_{k})$,
and calculated the performance
drops from $P(AP_{k+1},IP_{k})$ to these performances, i.e.,
$|P(AP_{k+1},IP_{k}) - P(AP_{k+2},IP_{k})|$,...,
$|P(AP_{k+1},IP_{k})-P(A_{4},IP_{k})|$, then these
performance drops were compared to the performance improvement.
The averaged performances (over 20 runs) of $AP_{k},..., AP_{4}$ on $IP_{k}$
are presented in Table~\ref{table2}.
The average performance improvement on $IP_{1}$
is $P(AP_{2},IP_{1})-P(AP_{1},IP_{1}) = 0.4321$,
and the two average performance drops on $IP_{1}$
are $P(AP_{3},IP_{1})-P(AP_{2},IP_{1}) = 0.0700$
and $P(AP_{4},IP_{1})-P(AP_{2},IP_{1}) = 0.1254$,
so the ratios between the performance drops
and the performance improvements on $IP_{1}$ are 16.20\% and 29.02\%.
Calculated in the same way,
the ratio on $IP_{2}$ is 17.98\%.
All the ratios between the performance drops
and the corresponding performance improvements
are below $30\%$.

\begin{table}[tbp]
  \footnotesize
  \centering
  \caption{Performances of $AP_{k}$,...$AP_{4}$ on $IP_{k}$.
  All the results are presented in terms
  of applicability, i.e., the proportion of the
  instances to which $AP$ is applicable, and are averaged over 20 runs.}
    \begin{tabular}{c c c c}
          \hline
          & \multicolumn{1}{c}{$IP_{1}$} & \multicolumn{1}{c}{$IP_{2}$} & \multicolumn{1}{c}{$IP_{3}$} \\
          \hline
    $AP_{1}$   & 0.3800 &          &  \\
    $AP_{2}$   & 0.8121 & 0.1456   &  \\
    $AP_{3}$   & 0.7421 & 0.6590   & 0.1735 \\
    $AP_{4}$   & 0.6867 & 0.5667   & 0.5654 \\
        \hline
    \end{tabular}
  \label{table2}
\end{table}

\begin{figure}
\centering
\includegraphics[scale = 0.6]{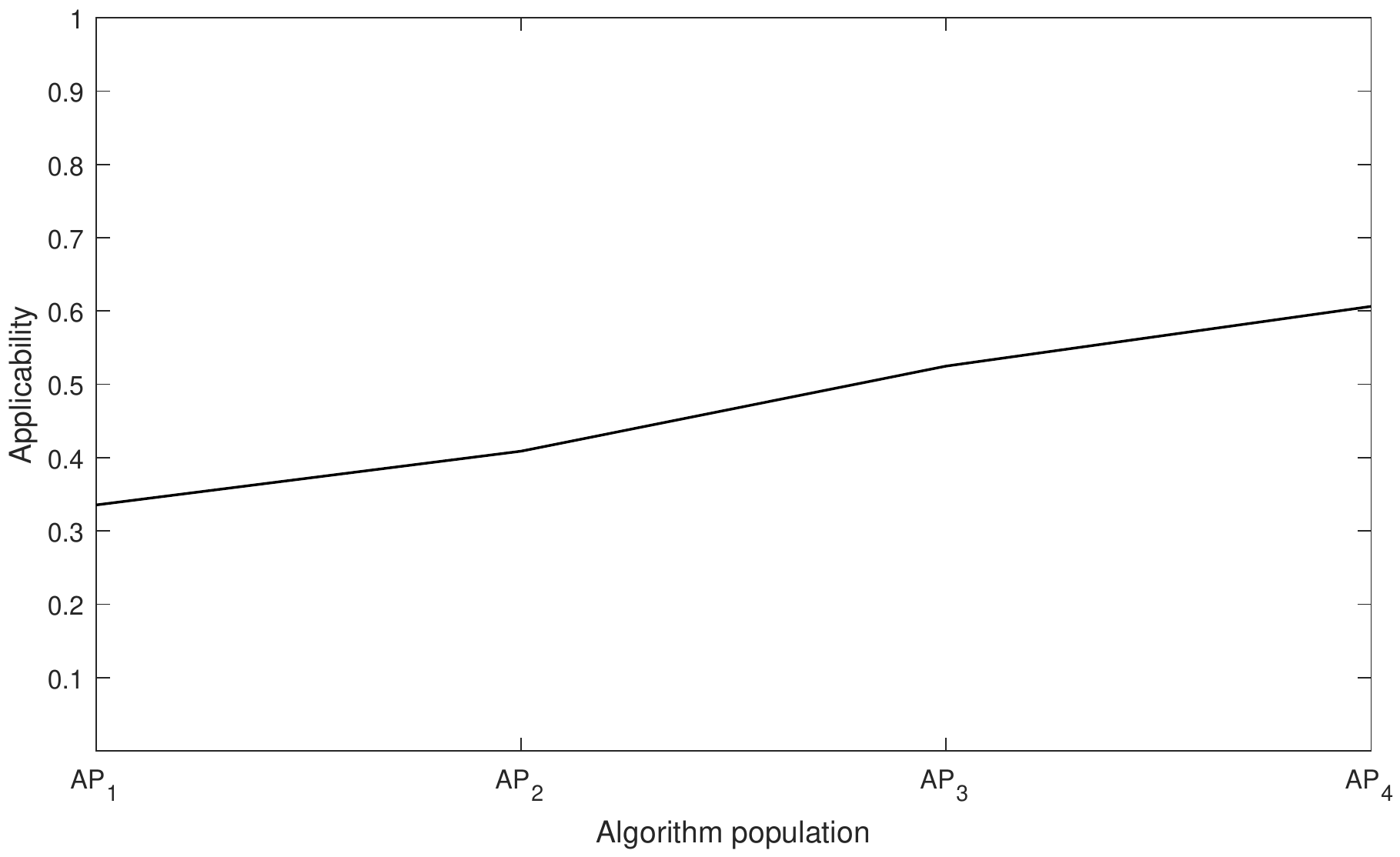}
\caption{The performances
of $AP_{1},AP_{2},AP_{3},AP_{4}$ on $IP_{training}$.
The performances are represented in terms
of applicability, i.e., the proportion of the instances to which AP is
applicable, and are averaged over 20 runs.}
\label{figure2}
\end{figure}

In order to check whether the performances of $AP$
on $IP_{training}$ are improved by LiangYi,
the algorithm populations obtained
from each cycle of LiangYi, i.e., $AP_{1},AP_{2},AP_{3},AP_{4}$,
were tested on $IP_{training}$.
The test results averaged over 20 runs are
depicted in Figure~\ref{figure2}. A constant improvement of the
performances of $AP_{k}$
on $IP_{training}$, according to the increase of $k$, is shown.

\subsubsection{Test Part}

\begin{figure}
\includegraphics[scale = 0.6]{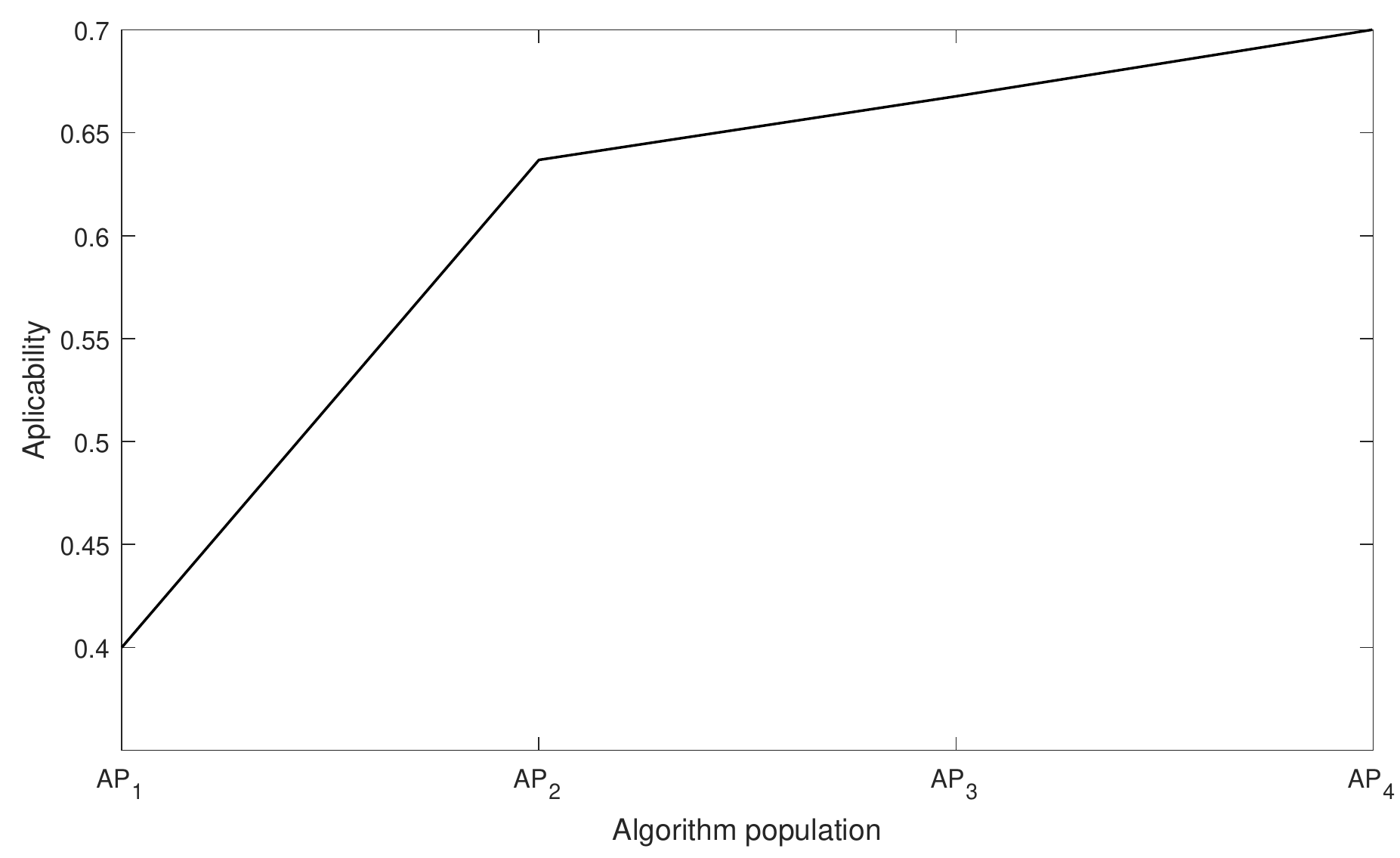}
\caption{The performances
of $AP_{1},AP_{2},AP_{3},AP_{4}$ on $IP_{test}$.
The performances are represented in terms
of applicability, i.e., the proportion of the instances to which AP is
applicable, and are averaged over 20 runs.}
\label{figure3}
\end{figure}

The algorithm population obtained
from each cycle of LiangYi,
i.e., $AP_{1},...,AP_{4}$
was tested on $IP_{test}$.
The test results averaged over 20 runs are depicted in
Figure~\ref{figure3}.
Once again, a constant improvement of the performances
of $AP_{k}$ on $IP_{test}$
according to the increase of $k$ is shown.

\section{Conclusion and Future Directions} 
\label{sec:conclusion}
This paper first put forward the concept of experience-based optimization (EBO)
which concerns improving solvers based on their past solving experience,
and summarized several previous research in a unified context,
i.e., offline training of EBO.
A new coevolutionary training method, dubbed LiangYi, was proposed. The most novel
feature of LiangYi is that, different from existing  methods,
it addresses selecting training instances and
training solvers simultaneously.
A specific instantiation of LiangYi on TSPs
was also proposed.
Empirical results showed
the advantages of LiangYi in comparison to ParHydra,
the state-of-the-art $APC$ method, on a huge test set
containing 10000 instances.
Moreover, through empirically investigating behaviours of
LiangYi, we confirmed that
LiangYi is able to continuously improve
the solver through training.


As discussed in the introduction, EBO is a far more broad direction
than merely offline training of problem solvers.
Further investigations may include:
\begin{enumerate}
\item Further improvements to LiangYi. Diversity preservation scheme, such
as speciation \cite{vcrepinvsek2013exploration} or negatively correlated search
\cite{tang2016negatively} can be introduced into LiangYi to explicitly promote
cooperation between different algorithms in $AP$.
Another tack is to use machine learning techniques to accelerate LiangYi.
Specifically, regression models and classification models can be used to
predict the performance of algorithms in $AP$ or instances in $IP$, without
actually evaluating them, which is very time-consuming.

\item Online mode of EBO.
Situations in which a solver faces a series
of different problem instances coming sequentially
pose new challenges.
For example, the objective in online mode
is to maximize the cumulative performance on all the instances.
Thus methods designed for this scenario must consider
making solvers perform well on current instances and
improving solvers for future instances
simultaneously.
Besides, in a dynamic environment,
the underlying properties of instances may change
overtime; therefore the solvers need to
keep detecting the changes of environment and
adapt to new instances.

\item Deeper understanding of the fundamental issues of EBO is also worthy of
exploration. For example, LiangYi actually maintains two adversary sets
competing against one another, which is a typical scenario where the game theory
can be applied. Besides, other more general issues in EBO include
the similarity measure between instances, a unified approach
to information extraction from solved instances, and theoretical proofs
of the usefulness of transmitting information between similar instances.


\end{enumerate}

\section{Acknowledgements}
This work was supported in part by the National Natural Science Foundation of China under Grant 61329302 and Grant 61672478; EPSRC under Grant EP/K001523/1 and EP/J017515/1, the Royal Society Newton Advanced Fellowship under Grant NA150123, and SUSTech. Xin Yao was also supported by a Royal Society Wolfson Research Merit Award.
\section*{References}

\bibliography{mybibfile}

\end{document}